\documentclass[lettersize,journal]{IEEEtran}

\usepackage{amsmath,amsfonts}
\usepackage{graphicx}
\usepackage{array}
\usepackage[caption=false,font=normalsize,labelfont=sf,textfont=sf]{subfig}
\usepackage{textcomp}
\usepackage{booktabs}
\usepackage{cite}
\usepackage{url}
\usepackage{hyperref}

\usepackage{algorithm}
\usepackage{algorithmic}

\usepackage{acronym} 

\acrodef{V2X}{Vehicle-to-Everything}
\acrodef{ACC}{Adaptive Cruise Control}
\acrodef{AV}{Autonomous Vehicle}
\acrodef{C-V2X}{Cellular V2X}
\acrodef{EM}{Emergency Message}
\acrodef{IVD}{Inter-Vehicle Distance}
\acrodef{RSS}{Responsibility-Sensitive Safety}
\acrodef{TTC}{Time to Collision}
\acrodef{URLLC}{Ultra-Reliable Low Latency Communication}
\acrodef{V2V}{Vehicle-to-Vehicle}
\acrodef{DRL}{Deep Reinforcement Learning}
\acrodef{RL}{Reinforcement Learning}
\acrodef{MDP}{Markov decision process}
\acrodef{MLP}{multilayer perceptron}
\acrodef{PPO}{Proximal Policy Optimization}
\acrodef{SAC}{Soft Actor-Critic}
\acrodef{BM}{Baseline Model}
\acrodef{E-V2X-BM}{Ethical-V2X-Baseline Model}

\begin{document}


\title{How to Brake? Ethical Emergency Braking with \\ Deep Reinforcement Learning}





\author{Jianbo Wang, Galina Sidorenko and Johan Thunberg~\IEEEmembership{Member,~IEEE}

\thanks{The authors are with the Department of Electrical and Information Technology (EIT), Lund University, SE-22100 Lund, Sweden (e-mail: ji0628wa-s@student.lu.se; galina.sidorenko@eit.lth.se; johan.thunberg@eit.lth.se).}
\thanks{This work was partially supported by the Wallenberg AI, Autonomous Systems and Software Program (WASP) funded by the Knut and Alice Wallenberg Foundation.}
}
%

\acrodef{V2X}{vehicle-to-everything}
\acrodef{ACC}{adaptive cruise control}
\acrodef{AV}{autonomous vehicle}
\acrodef{EM}{emergency message}
\acrodef{IVD}{inter-vehicle distance}
\acrodef{RSS}{Responsibility-Sensitive Safety}
\acrodef{TTC}{time-to-collision}
\acrodef{V2V}{vehicle-to-vehicle}
\acrodef{DRL}{Deep Reinforcement Learning}
\acrodef{MLP}{Multilayer perceptron}
\acrodef{MB}{Model based}

\maketitle


\begin{abstract}
Connected and automated vehicles (CAVs) have the potential to enhance driving safety, for example by enabling safe vehicle following and more efficient traffic scheduling. 
For such future deployments, safety requirements should be addressed, 
where the primary such are avoidance of vehicle collisions and substantial mitigating of harm when collisions are unavoidable.  
However, conservative worst-case-based control strategies come at the price of reduced flexibility and may compromise overall performance.
In light of this, we investigate how \ac{DRL} can be leveraged 
to improve safety in multi-vehicle-following scenarios involving emergency braking. 
Specifically, we investigate how \ac{DRL} with vehicle-to-vehicle communication can be used to ethically select an emergency breaking profile in scenarios where overall, or collective, three-vehicle harm reduction or collision avoidance shall be obtained instead of single-vehicle such.
As an algorithm, we provide a hybrid approach that combines \ac{DRL} with a previously published method based on analytical expressions for selecting optimal constant deceleration.
By combining \ac{DRL} with the previous method, the proposed hybrid approach increases the reliability compared to standalone \ac{DRL}, while achieving superior performance in terms of overall harm reduction and collision avoidance.

\end{abstract}


\section{Introduction}
\IEEEPARstart{A}{utonomous} vehicles (AVs) and connected such (CAVs) have been regarded a cornerstone of next-generation transportation systems, promising to reduce traffic accidents and fatalities through reliable automation strategies \cite{AVtraffic} \cite{TOURAN1999567}. Ensuring safety in multi-vehicle scenarios remains a critical challenge. Among various accident types, rear-end collisions, frequently caused by sudden deceleration of the leading vehicle, represent the most common and one of the most severe accident types \cite{rearend1}.

Numerous control strategies have been proposed for multi-vehicle coordination in these settings \cite{sidorenko2023ethical} \cite{geisslinger2023ethical} \cite{katrakazas2015real}. Notably, \cite{sidorenko2023ethical} introduced a longitudinal safety framework based on extending the \ac{RSS} model by explicitly calculating the minimum safe distance and the maximum safe communication latency/delay for specific scenarios. In \cite{sidorenko2024cooperation}, the benefit of \ac{V2X} communication in these settings is underscored; an algorithm for cooperative multi-vehicle safety is derived with a solution for minimum \acp{IVD}. 

The approaches and schemes mentioned above rely on conservative strategies for worst-case scenarios, which undermines their safety performance. While safety in theory is guaranteed, the required inter-vehicle distance is inflated, and due to insufficient modeling and inflexible decision making avoidance of avoidable collisions may not be ensured in practice. 

To overcome the limitations of the mentioned approaches in multi-vehicle scenarios, \acf{DRL} emerges as a promising tool.
By combining the decision-making framework of reinforcement learning with the function approximation capability of deep neural networks, \ac{DRL} can effectively operate in high-dimensional and continuous-control domains\cite{DRL1}.
Unlike classical methods, \ac{DRL} agents learn control policies through direct interaction with the environment, optimizing cumulative rewards that explicitly encode safety objectives (e.g., penalizing collision risks and encouraging minimal collision harm)\cite{DRL2} \cite{AVDRL}. 
This data-driven approach offers notable advantages, including the ability to handle non-convex and non-smooth Jacobian scenarios\cite{DRLcontrol1} .  

However, while prior studies~\cite{DRLcar1,DRLcontrol2,DRLcar2} have demonstrated the effectiveness of \ac{DRL} in various autonomous driving tasks (e.g., trajectory following, lane changing), its potential for improving \emph{multi-vehicle rear-end collision mitigation} remains largely underexplored.

This paper focuses on emergency braking for three-vehicle longitudinal driving scenarios, a common and representative scenario for safe following in real-world traffic \cite{longitudinal}\cite{longitudinal2}, and one that is closely associated with the occurrence of rear-end collisions \cite{rearend}. At the price of more coordination, three-vehicle interaction instead of pairwise interaction enables exploration of more degrees of freedom for improving safety.
To address such scenarios, \ac{V2X} communication is incorporated, with realistic communication delays and tight \ac{IVD} to ensure the fidelity of the proposed method~\cite{V2X}. We consider a critical emergency braking situation where the leading vehicle encounters an unforeseen obstacle, triggering maximum deceleration ($a_{\text{max}}$). This abrupt maneuver implies significant safety threats for the following vehicles due to severely reduced \ac{TTC} and control latency caused by \ac{V2X} communication delay/latency ($\tau$). Beyond the safety challenges, this setting also embodies an ethical problem, underscoring the need for control strategies that minimize overall harm rather than prioritizing a single vehicle~\cite{Ethical}.

Based on this three-vehicle scenario, we propose a \ac{DRL}-based approach that explicitly optimizes the ethical outcome expressed as overall harm. 
For rigorous performance evaluation, we use 
the method in~\cite{sidorenko2024cooperation}, referred to as \ac{E-V2X-BM}, as a benchmark. That method provides, via analytical expressions, overall harm as a function of system parameters for an introduced braking strategy using constant deceleration for Vehicle 2. 
Evaluation results show that compared to this baseline method, our proposed approach reduces overall collision harm and, for  certain \acp{IVD}, avoids collisions present for the \ac{E-V2X-BM}. 

As main algorithm, we develop a hybrid framework that integrates the learned \ac{DRL} policy with the \ac{E-V2X-BM}, leveraging \ac{V2X} communication to enhance safety reliability in multi-vehicle driving scenarios. This hybrid design establishes a guaranteed safety bound while improving the robustness and reliability of \ac{DRL} in practical deployment.

The contributions of this paper can be summarized as follows:
\begin{itemize}
    \item \textbf{Application of \ac{DRL} to Multi-Vehicle Longitudinal Safety:} 
We validate the applicability of the proposed \ac{DRL}-based approach to collision avoidance in multi-vehicle longitudinal driving scenarios, 
training policies with multiple representative \ac{DRL} algorithms.
Compared to the existing ethical baseline, the proposed approach enhances overall safety by avoiding collisions in cases where the baseline fails and mitigating the severity when collisions are unavoidable.

    \item \textbf{Hybrid DRL Framework for Reliability:} 
We propose a hybrid framework that integrates the \ac{E-V2X-BM} approach with \ac{DRL} to achieve improved reliability.
Within this framework, the proposed model serves as a guidance layer, steering decision-making toward balanced and reliable responses.
Experimental results show that the hybrid design not only enhances \ac{E-V2X-BM} performance but also achieves higher reliability compared to using \ac{DRL} alone.
\end{itemize}

The remainder of this paper is structured as follows: 
Section~\ref{sec:problem} presents the problem statement, including the multi-vehicle longitudinal driving scenario and key safety considerations. 
Section~\ref{sec:baseline_drl} describes the  \ac{DRL} framework and the proposed hybrid approach.
Section~\ref{sec:result} reports the experimental results including comprehensive evaluation metrics and analysis of the operational mechanisms behind safety improvements. 
Finally, Section~\ref{sec:conclusion} concludes the paper and discusses potential directions for future research.

\section{Problem Statement}
\label{sec:problem}



\begin{figure}[!t]
    \centering
    \includegraphics[width=1.0\linewidth]{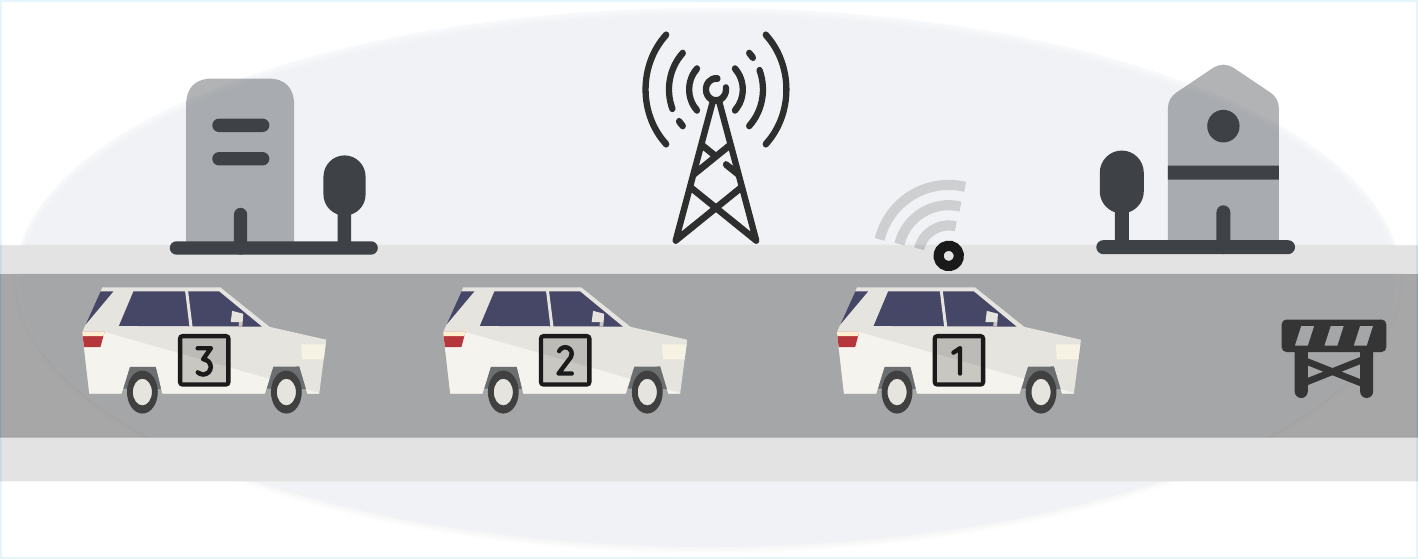}
    \caption{Illustration of the Three-Vehicle Longitudinal Driving Scenario with V2X Communication.}
    \label{fig:fig1}
\end{figure}

The main considered scenario is illustrated in Fig.~\ref{fig:fig1}. Three vehicles follow each other along a road. The leading vehicle is denoted Vehicle~1, the middle one Vehicle~2, and the last one Vehicle~3. We refer to $a_{l,i}^{\text{max}}$ as the maximum deceleration magnitude, also referred to as braking capacity, for Vehicle $i$. It is assumed that the vehicles communicate with each other via \ac{V2X} communication, enabling the sharing of relevant state, sensing, and emergency information. Without loss of generality, we assume that at time $t = 0$ Vehicle $1$ has to emergency brake (to \emph{e.g.}, avoid a potential collision with an unforeseen obstacle). Emergency braking in this context is modeled by Vehicle~1 decelerating with magnitude $a_{l,1}^{\max} > 0$. Due to latency and communication delays, Vehicle~2 starts braking at time $t = \tau_2 > 0$, while Vehicle~3 starts braking at time $t = \tau_3 > 0$. Vehicle~2 and Vehicle~3 can only brake in such a manner that their deceleration magnitudes do not exceed $a_{l,2}^{\text{max}} > 0$ and $a_{l,3}^{\text{max}} > 0$, respectively. The goal is to determine a feasible deceleration profile for Vehicle~2 that minimizes, and preferably avoids, collisions among the vehicles.

In this scenario, a simple braking strategy would be for Vehicle~2 and Vehicle~3 to apply their respective maximum braking capacities from time points $\tau_2$ and $\tau_3$, respectively, until full stop.
However, such a strategy would not necessarily be the least harmful one \cite{sidorenko2024cooperation}. It might result in a collision that could potentially have been avoided with a more sophisticated braking strategy.

This paper considers the situation where Vehicle~3 applies maximum deceleration while Vehicle~2 applies a braking strategy learned via \ac{DRL}. To be able to perform such a strategy, Vehicle~2 needs a small amount of information about Vehicle~1 and Vehicle~3 obtained via \ac{V2X} communication. This information includes the maximum braking capacities and the initial states. In general, the braking strategy of Vehicle~2 should be chosen to ensure that all three vehicles stop safely without collisions. However, in extreme situations, such as tight \acp{IVD}, high velocities and/or high latencies, the objective shifts to selecting a strategy that reduces the collision severity or harm as much as possible. 

In Section~\ref{sec:system_model} we provide the system model, including vehicle dynamics, collision modeling, and the definition of harm resulting from a collision. Then, in Section~\ref{sec:optimization_problem}, we formalize the problem outlined above by formulating it as a discrete-time optimization problem for Vehicle 2, whose solution may be mapped to continuous time via interpolation.

\subsection{System Model for emergency braking}\label{sec:system_model}

Let $x_i(t)$ and $v_i(t)$ denote the position and velocity, respectively, of Vehicle $i$ at time $t\in\mathbb{R}$, where $i \in \{1,2,3\}$. 
Since longitudinal motion is considered, the time evolution of the vehicles is modeled during collision-free intervals by:
\begin{equation}
\begin{bmatrix}
\dot{x}_i(t)\\
\dot{v}_i(t)
\end{bmatrix}
\!\!=\!\!
\begin{bmatrix}
v_i(t)\\
u_i\bigl(t\bigr)
\end{bmatrix}
\;-a_{l,i}^{\max}\!\leq \!u_i(t)\!\leq\! a_{u,i}^{\max}, \;i\in\{1,2,3\},
\label{dynamics}
\end{equation}
where the control signal $u_i(t)$ (longitudinal acceleration/deceleration) is bounded from below and above. 

At time $t = 0$, the leading vehicle (Vehicle~1) starts braking with maximum braking capacity until standstill:
\begin{equation}\label{eq:olle1}
u_1(t)= \begin{cases}
    -a_{l,1}^{\max}, & 0 \leq t \leq v_1(0)/a^{\max}_{l,1}, \\
    0, & t > v_1(0)/a^{\max}_{l,1}.
\end{cases} \quad 
\end{equation}

V2X communication introduces non-negligible latencies. We assume that Vehicle~3 has received and processed the message with the emergency braking information before the time-point \(\tau_3\) and starts braking at time \(t=\tau_3\) with maximum braking capacity until standstill. Hence
\begin{equation}\label{eq:olle2}
u_3(t)=
\begin{cases}
0,& t<\tau_3,\\
-a_{l,3}^{\max},& \tau_3 \leq t \leq \tau_3 + v_3(0)/a_{l,3}^{\max},\\
0, & t > \tau_3 + v_3(0)/a_{l,3}^{\max}. 
\end{cases}
\end{equation}

The middle vehicle (Vehicle 2) is a controllable agent that follows the control decision/policy $\pi_{\theta}(t)$ from time $\tau_2$ onward. The symbol $\theta$ captures parameters to be learned for the policy. The time $\tau_2$ is chosen sufficiently large to ensure that the necessary information for generating such control policy has been communicated from Vehicle 1 and Vehicle 3. This information comprises $a_{l,1}^{\max}$, $a_{l,3}^{\max}$, $\tau_3$, $v_1(0)$, $v_3(0)$, and relative distances $d_1(0) = x_1(0) - x_2(0)$, $d_2(0) = x_2(0) - x_3(0)$ (if they are not measured by onboard sensors of the ego vehicle). We also assume that Vehicle 2 is not involved in any collisions prior to time $\tau_2$. Thus, the control action of Vehicle 2 at emergency braking is given as
\begin{equation}
u_2(t)=
\begin{cases}
0,&t<\tau_2,\\
 \pi_\theta(\cdot),& t\ge \tau_2.
\end{cases}
\end{equation}
where $\pi_\theta(\cdot)$ is the deterministic policy, mapping the environment state to a control action in order to avoid collisions or mitigate their severity.

For $t > \tau_2$,
the control input of Vehicle~2 must satisfy
\begin{equation}\label{eq:4}
-\,a_{l,2}^{\max} \;\le\; u_2(t) \;\le\; a_{u,2}^{\max},
\end{equation}\label{eq:5}
Furthermore, for $t > \tau_2$, the velocity shall satisfy   
\begin{equation}\label{eq:52}
0\le\; v_2(t) \;\le\; v_2^{\max} > 0,
\end{equation}
which means that reverse motion is not possible and that the velocity is subject to an upper bound. 

The relative distance between consecutive vehicles, 
\begin{equation}\label{eq:rel_distance}
d_i(t)=x_i(t)-x_{i + 1}(t), \quad i \in \{1,2\}
\end{equation}
is independent of the choice of global coordinate system and is used to define safety. Collision between Vehicle $i$ and Vehicle $i+1$ implies that $d_i(t)$ becomes non-positive ($d_i(t) \le 0$) for $i \in \{1,2\}$. We define collision indicator function as follows
\begin{align} \label{gamma}
\Gamma(d_i) & = 
\begin{cases}
1, & \text{if } d_i \leq  0 ,\\
0, & \text{otherwise},
\end{cases}, \quad i \in \{1,2\}, 
\end{align}
where at time $t$, $\Gamma(d_i(t)) = 1$ indicates that a collision has occurred between Vehicle $i$ and Vehicle $i+1$ for $i \in \{1,2\}.$
For the special case where $d_i(t) = 0$ and $v_i(t) - v_{i+1}(t) = 0$, 
the vehicles are merely touching without any energy exchange. As the present study focuses on collision harm, such a touching scenario with zero relative velocity is considered to have no harm contribution and is therefore not counted as a collision.

 A variety of metrics have been proposed for quantifying collision severity, from momentum exchange to detailed structural models \cite{Harm,geisslinger2023ethical}. Empirical studies have demonstrated that the relative velocity between the two vehicles at the moment of collision is strongly correlated with the severity of the driver's injury \cite{Harm2,Harm3}. We define the relative velocity at time $t$ as
\begin{equation}\label{eq:rel_velocity}
\tilde v_i(t)=v_{i+1}(t)-v_i(t), \quad i \in \{1,2\}. 
\end{equation}
At the collision time $t_c$, we denote $\tilde v_i^-(t_c)$ and $\tilde v_i^+(t_c)$ as the relative velocities immediately before and after the collision, respectively. 

In \cite{sidorenko2023ethical,sidorenko2024cooperation,geisslinger2023ethical} the relative velocity before the collision and the masses of the vehicles are combined to calculate a so-called harm coefficient. Here we use an energy‑equivalent expression by using the square of the relative velocity. 
The harm $H_i(d_i, \tilde v_i)$ for Vehicle $i$, involved in a collision with Vehicle~$i+1$ is defined as 
\begin{equation}\label{eq:harm}
H_i(d_i, \tilde v_i) = \left (\frac{m_{i+1}}{m_i + m_{i+1}} \tilde v_i^2 \right) \cdot \Gamma(d_i).
\end{equation}
Similarly, the harm to Vehicle~$i+1$ involved in a collision with Vehicle~$i$ is obtained as
\begin{equation}\label{eq:harm_i+1}
    H_{i+1}(d_i, \tilde v_i) = \frac{m_i}{m_{i+1}} \, H_i(d_i, \tilde v_i),
\end{equation}
where the mass ratio term accounts for the distribution of kinetic energy between the two vehicles. With this notation, $H_2(d_1, \tilde v_1)$ is the harm for Vehicle 2 in a collision with Vehicle~1, whereas $H_2(d_2, \tilde v_2)$ 
is the harm for Vehicle~2 resulting from a collision with Vehicle~3. 

To improve physical fidelity, the instantaneous change in velocities upon impact is modeled using the principle of conservation of momentum for the two vehicles. For a rear-end collision between Vehicle~\(i\) and Vehicle~\(i+1\), the immediate after-collision velocities $v_{i}^{+}$ and $v_{i+1}^{+}$ satisfy
\begin{equation}\label{eq:11}
m_i v_i^{+} + m_{i+1} v_{i+1}^{+} = m_i v_i^{-} + m_{i+1} v_{i+1}^{-},
\end{equation}
where $v_i^{-}$ denotes the velocity of Vehicle~$i$ before the collision for $i \in \{1,2\}$. \\
Energy dissipation during the collision is modeled by using a coefficient of restitution $e \in [0, 1]$:
\begin{equation}\label{eq:12}
e = \frac{v_{i}^{+} - v_{i+1}^{+}}{v_{i+1}^{-} - v_{i}^{-}}. 
\end{equation}
Thus, $e$ is defined as the ratio of the relative velocity immediately after the collision to that before the collision. 
Its magnitude reflects the amount of energy dissipated during impact and depends on material properties, structural stiffness, and impact geometry. 
Empirical studies indicate that typical rear-end collisions correspond to $e \leq 0.3$~\cite{Harm-e}.

From \eqref{eq:11} and \eqref{eq:12}, the following velocity exchange law is derived.
\begin{equation}\label{eq:13}
\begin{aligned}
v_i^{+} &= v_i^{-}+ \frac{m_{i+1}}{m_{i+1}+m_i}\bigl(1+e\bigr)\tilde v_i^-
\\v_{i+1}^{+} &= v_{i+1}^{-}- \frac{m_i}{m_{i+1}+m_i}\bigl(1+e\bigr)\tilde v_i^-.
\end{aligned}
\end{equation}
The formulation \eqref{eq:13} provides an update rule for subsequent motion planning and collision severity assessment. As such, the dynamics of the system (which is partly described by \eqref{dynamics}) is hybrid, where velocities jump to new values when relative positions are zero and relative velocities are positive. 

\subsection{Optimization Problem }\label{sec:optimization_problem}


To implement the \ac{DRL}-strategy, we first discretize the dynamics of Vehicle~2. We then formulate an optimization problem, which incorporates the described model along with input constrains \eqref{eq:4} and state constraints \eqref{eq:52}.  

We use a zero-order hold discretization of \eqref{dynamics} (see \eqref{dynamics2} below), where the sampling period is $\Delta t$ and the continuous time signal is held constant between consecutive sampling instants. The time discretization begins at time $t = \tau_2$, thus discrete time instant $n$ corresponds to the continuous time instant $t_n=\tau_2 + n \Delta t$. Furthermore, to capture the hybrid nature of the system, where velocities jump to new values at vehicle collision according to \eqref{eq:13}, the additional function $\tilde f$ is used. This function comprises the composition of the two functions $\tilde f_1$ and $\tilde f_3$, which capture collisions of Vehicle~2 with Vehicle~1 and Vehicle~3, respectively. 
\begin{equation}
\underbrace{
\begin{bmatrix}
{x_2(n+1)}\\
{v_2(n+1)}
\end{bmatrix}}_{s_2(n+1)}
=
\underbrace{
\begin{bmatrix}
 1 & \Delta t\\
 0 & 1
\end{bmatrix}
\begin{bmatrix}
 x_2(n)\\
 \tilde f(n,s_2(n))
\end{bmatrix} + 
\begin{bmatrix}
 \frac{(\Delta t)^2}{2}\\
\Delta t
\end{bmatrix} u_2(n)}_{f(s_2(n), u_2(n)},
\label{dynamics2}
\end{equation}
where $s_2(n)=[x_2(n),v_2(n)]^\top$ denotes the state of Vehicle~2 at discrete time index $n \geq 0$ and 
\begin{equation}
\begin{aligned}
\tilde f(n, s_2(n)) & =\tilde f_1(n, [x_2(n), \tilde f_3(n, s_2(n))]^{\top}),
\end{aligned}
\end{equation}
where
\begin{equation}
\begin{aligned}
\tilde f_1(n, s_2(n)) & = v_{2}(n) - \frac{m_1(1+e)}{m_{2}+m_1}\tilde v_{1}(n)\Gamma(d_1(n)), \\
\tilde f_3(n, s_2(n)) & = v_{2}(n) + \frac{m_3(1+e)}{m_{2}+m_3}\tilde v_{2}(n)\Gamma(d_2(n)).
\end{aligned}
\end{equation}
Here, $d_1(n)$ and $d_2(n)$ are defined by \eqref{eq:rel_distance}, $\tilde v_1(n)$ and $\tilde v_2(n)$ are defined by \eqref{eq:rel_velocity} with discrete time step $n$ corresponding to $t_n=\tau_2+n\Delta t$.
The velocities of Vehicle~1 and Vehicle~3 are updated in an equivalent fashion when collisions with Vehicle~2 occur. Besides these events, Vehicle~1 and Vehicle~3 brake with constant deceleration (from time $\tau_3$ for Vehicle~3) until standstill. Thus, their positions and velocities are regarded as functions of $n$ and $s_2$. Hence the updating rules at collisions are given as functions of $n$ and $s_2$ only. For brevity, we do not provide the explicit discrete-time expressions for the positions and velocities of Vehicle 1 and Vehicle 3 as functions of $n$ and $s_2$; these follow directly from \eqref{dynamics}, \eqref{eq:olle1} or \eqref{eq:olle2}, and \eqref{eq:13} together with discretization. 

The objective function in the finite-horizon optimization problem with $N$ time-steps comprises a sum of $N$ terms 
\begin{equation}
    J = \sum_{n = 0}^{N-1}J_n.
\end{equation}
Each term $J_n$ itself consists of two components
\begin{equation}\label{eq:total_objective}
\begin{aligned}
J_n \;=\;& J_{\text{\text{jerk}},n}  + J_{\text{\text{harm}},n},
\end{aligned}
\end{equation}
where both $J_{\text{\text{jerk}},n}$ and $J_{\text{harm},n}$ are functions of the state $s_2$ and the control input $u_2$. For brevity, the explicit dependence on $s_2$ and $u_2$ is omitted in the provided definitions.

The first term, $J_{\text{\text{jerk}},n}$, promotes control smoothness and is defined as
\begin{align}
J_{\text{jerk},n} &= \left( \frac{u_2(n) - u_2(n-1)}{\Delta t} \right)^2.
\end{align}
The second component, $J_{\text{harm},n}$, penalizes severity of a collision when it is unavoidable, and is defined as
\begin{equation}\label{eq:Jharmn}
\begin{aligned}
J_{\text{harm},n} =~& H_1(d_1(n), \tilde v_1(n))  +   H_2(d_1(n), \tilde v_1(n))\\
&+ H_2(d_2(n), \tilde v_2(n)) +   H_3(d_2(n), \tilde v_2(n)),
\end{aligned}
\end{equation}
where $H_i$ for $i \in \{1,2,3\}$ is defined by \eqref{eq:harm} and \eqref{eq:harm_i+1}.  $J_{\text{harm},n}$ remains zero during safe driving but jumps to a (large) positive value if a collision occurs.

Finally, the finite-horizon, discrete-time, constrained optimal control problem over $N$ time steps is formulated below
\begin{equation} \label{Jfunc}
    \begin{cases}
     ~    \underset{\tilde u}{\text{minimize}} ~J(\tilde u, \tilde{s}) =  \sum\limits_{n=0}^{N-1}
    {J}_n(\tilde u, \tilde{s}) \\
     \hspace{1.cm}\text{s.t. } s_2(n) =f(s_2(n-1),u_2(n-1)), \\
     \hspace{3cm} \text{for }n \in \{1,2,\ldots,N-1\}, \\
     \hspace{1.6cm} g(\tilde s,\tilde u)\le0, \\
     \hspace{1.6cm} \tilde u = [u_2(0), u_2(1), \ldots, u_2(N-1)], \\
     \hspace{1.6cm} \tilde s = [s_2(0), s_2(1), \ldots, s_2(N-1)].
    \end{cases}
\end{equation}
Here, function $g$ captures state and input constraints provided in \eqref{eq:4} and \eqref{eq:52}.

The objective function 
in the optimization problem~\eqref{Jfunc}
exhibits a non-convex and non-smooth Jacobian (due to collisions), making the application of traditional gradient-based optimizers computationally challenging.
Therefore, we propose an adaptive \ac{MDP} scheme  based on the optimization problem~\eqref{Jfunc} where the solution is obtained by using \ac{DRL}. After this, we introduce a hybrid approach that combines \ac{DRL} with a conservative control algorithm \ac{E-V2X-BM}~\cite{sidorenko2024cooperation}, yielding a policy that is both safer and ethically aligned.



\section{\acf{DRL} and Hybrid DRL}
\label{sec:baseline_drl}

In this section, we first introduce the essential concepts of \ac{DRL} needed. We then present the learning algorithms used and subsequently discuss the detailed design of the \ac{RL} framework, including state space and reward shaping. 
Next, the hybrid method is proposed that combines the DRL approach with \ac{E-V2X-BM}~\cite{sidorenko2024cooperation}. The purpose of the hybrid approach is to provide a runtime safety shield on the learned policy, delivering a safer and more reliable control strategy. The section ends with a description of the network architecture. 

\subsection{Deep Reinforcement Learning}
\Ac{DRL} employs neural networks as function approximators and learns optimal policies through a closed-loop process of trial-and-error interactions guided by reward feedback. These characteristics make \ac{DRL} well-suited for multi-vehicle safe following problems~\cite{AVDRL2}~\cite{DRLdriving}. In practice, \ac{DRL} policies are often realized using compact neural architectures -- typically two- to four-layer \acp{MLP} -- which incur low computational overhead. The networks directly output control actions, providing a one-to-one mapping that is highly suitable for embedded hardware. Empirical studies~\cite{HardwareDRL2} \cite{hardwareDRL1} report inference latencies that satisfy vehicular control requirements, while hardware-oriented analyses~\cite{HardwareDRLdelay} further confirm the feasibility of deploying such \ac{DRL} controllers on resource-constrained devices.

Nevertheless, \ac{DRL} suffers from a well-documented drawback: the black-box nature of its policies can undermine reliability in practical scenarios \cite{DRLblackbox}. To overcome this limitation while preserving the practical advantages of \ac{DRL}, we propose a hybrid approach that merges \ac{DRL} with \ac{E-V2X-BM}~\cite{sidorenko2024cooperation} that provides safety bounds through analytical expressions. This integration yields a robust and trustworthy braking control policy suitable for safety-critical vehicle-following scenarios.

\subsection{DRL Algorithms Used}\label{sec:alg:used}
We use two \ac{DRL} algorithms, \Ac{PPO}~\cite{PPO} and \Ac{SAC}~\cite{SAC1}, briefly explained below. In Section~\ref{sec:result} we later compare the two choices of \ac{DRL} algorithms. 

\Ac{PPO}~\cite{PPO} is an on-policy reinforcement learning algorithm designed to improve policy performance while avoiding destabilizing with  excessively large updates. 
It uses the importance sampling ratio and augments the standard policy-gradient objective with a clipping mechanism.
The importance sampling ratio $\rho_{n}(\theta)$ is defined as:
\begin{equation}
\rho_{n}(\theta) =
\frac{\pi_{\theta}(a_2(n) \mid s(n))}
     {\pi_{\theta_{\mathrm{old}}}(a_2(n) \mid s(n))}
\label{eq:ratio}
\end{equation}
where $\theta_{\mathrm{old}}$ and $\theta$ denote the parameters of the old (previous) and current policies, respectively; $a_2(n)$ and $s(n)$ is the action and the state, respectilvely, at timestep $n$. $\pi_{\theta}(a_2(n) \mid s(n))$ is a probability of action $a_2(n)$ under the current policy while $\pi_{\theta_{\mathrm{old}}}(a_2(n) \mid s(n))$ denotes probability under the old policy. Further below, to save space, we denote the state at time step \(n\) by \(s_n\) instead of \(s(n)\).

Clipping of $\rho_n(\theta)$ between $1-\varepsilon$ and $1+\varepsilon$
limits the maximum change between the old policy and the updated policy.
The clipped objective is given by
\begin{equation}
\scalebox{0.9}{$
\mathcal{L}^{\mathrm{CLIP}}(\theta)=
\mathbb{E}\!\Bigl[
\min\!\bigl(
      \rho_{n}(\theta)\,\hat{A}_{n},\;
      \mathrm{clip}(\rho_{n}(\theta),1-\varepsilon,1+\varepsilon)\hat{A}_{n}
\bigr)
\Bigr]
\label{eq:ppo_clip}
$}
\end{equation}
where $\varepsilon$ is the hyper parameter that controls the permissible deviation per update, and  $\hat{A}_{n}$ is the estimated advantage defined as:
\begin{equation}
\hat{A}_n
= \sum_{l=0}^{\infty} (\gamma \lambda)^{l}\,\delta_{n+l},\;
\delta_n = r_n + \gamma\,V_\theta(s_{n+1}) - V_\theta(s_n)
\end{equation}
where $\gamma$ is the discount factor, $r_n$ is the reward at time instance $n$, $\lambda$ is GAE parameter that keeps the balance of bias and variance, $V_\theta(s_n)$ is the value function corresponding to the state $s_n$ at time step $n$, and $\delta_n$ represents the temporal-difference error. 

Finally, the complete \ac{PPO} loss function is given by:
\begin{equation}
\scalebox{0.89}{$
\mathcal{L}^{\text{PPO}}(\theta)
= \mathcal{L}^{\text{CLIP}}(\theta)
- c_1\,\mathbb{E}\!\bigl[(V_\theta(s_n)-V_{\text{target}})^2\bigr]
+ c_2\,\mathbb{E}\!\bigl[\mathcal{H}(\pi_\theta)\bigr]
$}
\end{equation}
where $c_1$ and $c_2$ are loss coefficients, $\mathcal{H}(\pi_\theta)$ denotes the policy entropy, and $V_{\text{target}}$ is the target value. By maximizing the $\mathcal{L}^{\text{PPO}}$ in each iteration, \ac{PPO} achieves stable updates, encourages exploration through the entropy term, and exhibits good convergence performance.

\Ac{SAC}, proposed in~\cite{SAC1} and further developed in~\cite{SAC2}, is an off-policy, entropy-regularized reinforcement learning algorithm that jointly maximizes the expected return and policy entropy. It employs dual Q-critic architecture and a stochastic policy and an adaptive temperature parameter to maximize the expected return while maintaining sufficient exploration. The optimal policy is given by
\begin{equation}
\pi^{*} = \arg \max_{\pi} \sum_{t} 
\mathbb{E}_{(s_n, a_n) \sim \pi}
\Big[ r_n + \alpha \mathcal{H}\big(\pi(\cdot \mid s_n)\big) \Big]
\end{equation}
where $\mathcal{H}(\pi(\cdot|s_{n})) = -\mathbb{E}_{a\sim\pi}[\log\pi(a|s_{n})]$ 
denotes the policy entropy, and $\alpha>0$ is a temperature coefficient 
that balances exploration and exploitation. 

In the follow-up work~\cite{SAC2}, to automatically adjust the temperature parameter $\alpha$ is treated as a learnable parameter, updated via entropy-constrained optimization:

\begin{equation}
\label{eq:alpha_loss}
\alpha_n^* = \arg\min_{\alpha_n} \; \mathbb{E}_{a_n \sim \pi_n^*} 
\Big[ -\alpha_n \log \pi_n^*(a_n \mid s_n, \alpha_n) - \alpha_n \bar{\mathcal{H}} \Big]
\end{equation}
where $\bar{\mathcal{H}}$ is a predefined target entropy. 

Entropy regularization, combined with the twin-critic design, provides SAC with stable exploration and robust convergence, making it particularly effective for high-dimensional, safety-control tasks considered in this work.

\subsection{MDP formulation}
The three-vehicle following scenario is abstracted as an \ac{MDP} where essential components are described below.

\noindent
\textbf{State:} the system state or observation $s(n)$	
comprises the (deterministic) actions of other vehicles and positions and velocities of all vehicles.   

\noindent 
\textbf{Action:}
in the problem considered, the \ac{RL} agent controls only Vehicle~2. At each time step, the policy outputs
\begin{equation}
  a_2 \in [-{a}_{l,2}^{\max},\, {a}_{u,2}^{\max}] \subset \mathbb{R},
\end{equation}
where the scalar action $a_2$ is the longitudinal acceleration/deceleration applied to the ego vehicle. The actions $a_1$ and $a_3$ are deterministic and given by~\eqref{eq:olle1} and \eqref{eq:olle2}, respectively. 

\noindent 
\textbf{Transition:} $\mathcal{P}(s({n+1})|s(n),a_2(n))$ captures the dynamics \eqref{dynamics2}. 

\noindent 
\textbf{Reward:}
Reward shaping is done by modifying $J(\tilde u, \tilde s)$ in \eqref{Jfunc} and reversing its sign so that maximizing the return corresponds to minimizing the objective. Additional shaping terms are introduced to facilitate learning. All-in-all, the reward comprises four components that are added together. These terms are explained below. 

\textit{Collision-severity penalty}: this term, which corresponds to $J_{\text{harm},n}$ \eqref{eq:Jharmn}, penalizes collisions by generating a large cost proportional to the relative velocity:
\begin{equation}
r_{\text{collision}} 
= -\sum_{i=1}^{2}\,k_{\text{energy,i}} \bigl(\tilde v_i(n)\bigr)^{2}\;
\Gamma(d_i(n)),
\end{equation}
where $\tilde v_i(n) $ is defined by \eqref{eq:rel_velocity}, and $k_{\text{energy,i}}$ are weight coefficients for $i \in \{1,2\}$ that balance the importance of this penalty relative to other reward components.

\textit{Collision-risk penalty}: to explicitly capture collision risk and provide tense penalties during training, both from the leading and following vehicles, we adopt a \ac{TTC}-based penalty in combination with a distance-based shaping term.  
Given the relative distance $d_i$ and relative velocity $\tilde v_i$, 
the instantaneous collision risk is defined as:
\begin{equation}
r_{\text{risk}} = \sum_{i=1}^{2}\Bigr( -\sigma\!\left(- \dfrac{\mathrm{TTC}_i}{\tau} \right) + \; \underbrace{-\sigma\!\bigl(k_d  (d_{target}- d_i\bigr))}_{\text{Distance penalty}}\Bigr),
\end{equation}
where 
$\mathrm{TTC}_i = \frac{d_i - d_{\mathrm{safe}}}{\tilde v_i}$
is the estimated time to collision, $\tilde v_i$ is defined by \ref{eq:rel_velocity}, $k_d > 0$ is a scaling coefficient, $d_{\mathrm{safe}}\geq 0$ is the safety buffer, $d_{target}\geq 0$ is the target safe distance, and $\tau > 0$ acts as a time-scaling factor in the \ac{TTC} term. 
Here, $\sigma$ denotes the sigmoid function $\sigma(x) = \frac{1}{1 + e^{-x}}$. It is introduced to bound the risk terms within a finite range, ensuring numerical stability during training and providing smooth gradients that facilitate policy optimization \cite{Sigmoid}.


\textit{Control-smoothness penalty}: this term, corresponding to  $J_{\text{\text{jerk}},n}$, aims to avoid overly aggressive control input:
\begin{equation}
r_{\text{jerk}}
\;=\;
- J_{\text{jerk},n}
\end{equation}

\textit{Terminal reward}: to encourage safe completion of an episode, a sparse constant reward is granted whenever the episode terminates without any collisions:
\begin{equation}
r_{\text{terminal}}=
\begin{cases}
R_{\text{safe}}, & \text{if no collision} \\
0, & \text{otherwise}.
\end{cases}
\end{equation}

\noindent Therefore, the total reward is defined as:
\begin{equation}
r_{\text{total}}
\;=\; w_h r_{\text{collision}} + w_p r_{\text{risk}} + w_j r_{\text{jerk}} + r_{\text{terminal}}
\end{equation}
where $w_h$, $w_p$, and $w_j$ are nonnegative weighting coefficients that balance 
collision avoidance, risk sensitivity, and motion smoothness, respectively.

\subsection{Hybrid DRL with E-V2X-BM}
In~\cite{sidorenko2023ethical}, the method we refer to as \ac{E-V2X-BM} was proposed for assessing and improving safety in emergency braking for the three-vehicle following scenario. This method is based on analytic expressions. The setting in~\cite{sidorenko2023ethical} was similar to the one considered in this paper, but the deceleration of Vehicle 2 was assumed to be constant and did not vary during braking. 

In \ac{E-V2X-BM}, analytical expressions are used to implement an algorithm that allows to calculate the so-called expected harm $H_{\text{total}}$ for all three vehicles, depending on the control input of Vehicle 2. This expected harm is equal to the overall harm or total harm we consider in this paper. We will henceforth simply refer to it as harm. The provided analytical functions depend on the initial \acp{IVD} $d_i(0)$, velocities, $v_i(0)$, deceleration capabilities  $a_{l,i}^{\max}$, and the \ac{V2X} delays $\tau_i$ for all $i$. Given the calculated ${H}_{\text{total}}$, the minimum of the harm can be found, $H_{*}=\min_{a_2} {H}_{\text{total}}(a_2)$, as well as the corresponding deceleration magnitude \(a_{*}\):
\begin{equation} \label{Eq:BaselineHarm}
a_{*} = \arg\min_{a_2} {H}_{\text{total}}(a_2).
\end{equation}
In scenarios where collisions can be avoided, i.e., $H_*=0$, the corresponding constant deceleration is possible to choose from an interval. That is, any constant deceleration $a_*$ within this interval guarantees a collision-free stop. When no constant deceleration results in a collision-free scenario, $a_*$ is uniquely determined as the deceleration that minimizes harm.





We use $H_{*}$ as a harm baseline which is used as a benchmark against \ac{DRL}-based strategies. 
Furthermore, $H_{*}$ is used as a safety threshold in the proposed hybrid approach. Any \ac{DRL} output with harm exceeding this threshold is automatically rejected. Compared to using reinforcement learning only, this provides a lightweight mechanism that is provably safe, transparent, and reliable; a hybrid control framework that integrates \ac{DRL} with \ac{E-V2X-BM} to achieve high performance with reliability.

The hybrid algorithm includes two essential components: 
\begin{enumerate}
\item \emph{Safety Bound}. 
When the prediction horizon contains no imminent collision, the DRL output $a_{\mathrm{RL}}$ is executed directly. If an emergency is expected, the harm $H_*$ is computed with the \ac{E-V2X-BM} approach. 

\item \emph{Actor Layer}. 
The actor $\pi_{\mathrm{RL}}$ is implemented as a two-layer ReLU network that outputs a control command $u_2$ at each timestep. If the prediction horizon contains no imminent collision, $u_2$ is executed directly.
\end{enumerate}

The two components operate as follows: once an emergency is detected, the algorithm compares the predicted harm of the actor’s output with that of \ac{E-V2X-BM}, i.e., $H_*$, and selects the strategy with the lowest harm as the final strategy. 
Specifically, upon receiving an emergency braking alert, Vehicle 2 immediately executes Algorithm 1. First, the safety bound $H_*$ is calculated, with the use of real-time \ac{V2X} measurements (headway, velocities, and communication delays). Then, the harm $H_{\mathrm{RL}}$ for the \ac{DRL} policy is computed. The result is then compared against the safety bound $H_{*}$ to produce the safe decision $\beta_{\text{safe}}$, where $\beta_{\text{safe}}{=}1$ corresponds to selecting the \ac{DRL} policy and $\beta_{\text{safe}}{=}0$ to selecting the \ac{E-V2X-BM} baseline. 


\begin{algorithm}[t]
\renewcommand{\algorithmicrequire}{\textbf{Input:}}
\renewcommand{\algorithmicensure}{\textbf{Output:}}
\caption{Hybrid method}
\label{alg:hybrid}
\begin{algorithmic}[1]   
\REQUIRE State $\mathbf{s}(n)$. 
\ENSURE Control sequence $\{u_2(n)\}$
\STATE Compute Safety baseline $(a_{*}, H_{*})$ by Eq.~\ref{Eq:BaselineHarm}
\STATE Set initial RL command $a_{RL} \leftarrow \pi_{\mathrm{RL}}(\mathbf{s}(n))$
\STATE Set expected harm $H_{\mathrm{RL}} \leftarrow 0$

\FOR{$n = 1$ to $N$}
    \STATE Predict next state under current RL action $a_{RL}$: \\$\mathbf{s}' \leftarrow \mathcal{P}(\mathbf{s}(n), a_{\mathrm{RL}})$
    \STATE Update RL command for the predicted state: \\$a_{\mathrm{RL}} \leftarrow \pi_{\mathrm{RL}}(\mathbf{s}')$
    \IF{\textsc{Collision}}
        \STATE Compute step harm \\$H_n \leftarrow H(s')$ by Eq.~\eqref{eq:Jharmn}
    \ENDIF
    \STATE Update expected harm: \\$H_{RL} =  H_{RL} +H_n$
\ENDFOR
\STATE \textbf{Decision:}
\IF{$H_{\mathrm{RL}} \le H_{\mathrm{*}}$}
    \STATE Select DRL strategy: $\beta_{\text{safe}} \leftarrow \ 1$
\ELSE
    \STATE Select E-V2X-BM strategy: $\beta_{\text{safe}} \leftarrow \ 0$
\ENDIF

\RETURN Based on $\beta_{\text{safe}}$, select the strategy to generate the control sequence $\{u_2(n)\}$.

\end{algorithmic}
\end{algorithm}

\subsection{Network Architecture}
In this paper, a lightweight multilayer perceptron is used to train the policy \(\pi_\theta\). Specifically, the policy network consists of two fully connected hidden layers, each with a width of 256 nodes, followed by a ReLU activation function.
In \ac{PPO} framework, the value network $V_{\phi}(s)$ that is trained to approximate value function has an additional hidden layer with 128 nodes to enhance its ability to approximate higher-order value functions.
In \ac{SAC} framework, both the policy network and the double Q network adopt a symmetrical structure: two fully connected hidden layers, each with 256 nodes, using ReLU activation function.

\section{Experimental Results and Observations}
\label{sec:result}
This section first establishes the feasibility of applying \ac{DRL} to the three-vehicle safety scenarios considered. Then, the effectiveness and reliability of the proposed hybrid method is validated. Finally, multi-episode simulations demonstrate that \ac{DRL}-based approaches reduce collision harm and improve safe car-following performance compared with \ac{E-V2X-BM}.

\subsection{Experimental Set-Up}
A three-vehicle emergency braking simulation platform is used for training and evaluation. The two different \ac{DRL} algorithms described in Section~\ref{sec:alg:used}, \ac{SAC} and \ac{PPO}, are used for policy learning with hyperparameters summarized in Table~\ref{table:params}. In the evaluation, the proposed \ac{DRL}-based approaches and the hybrid safety method are compared against two baselines: the \ac{E-V2X-BM} method and a non-ethical driving policy where Vehicle~2 applies maximum possible deceleration during braking to increase the distance from Vehicle~1 in front. Given the primary emphasis on safety, collision harm calculated by~\eqref{eq:harm} is selected as the principal performance metric. For fair comparison, we follow the protocol presented in~\cite{sidorenko2023ethical} for the  \ac{E-V2X-BM} method, where the harm associated with each pairwise collision is counted only once.

To facilitate reproducibility and further research, the complete environment configuration, including vehicle dynamics configurations, parameter weights and the detailed reward-function design, will be available on \href{https://github.com/S11Cho/Ethical-Vehicle-RL}{GitHub}.

\begin{table}[t]
\centering
\caption{PPO and SAC hyperparameters}\label{table:params}
\label{tab:ppo-sac}
\renewcommand{\arraystretch}{1.15}
\begin{tabular}{lll}
\hline
Parameter & Algorithm & Value \\
\hline
Policy & PPO/SAC & \texttt{MlpPolicy} \\
Learning rate & PPO/SAC & \texttt{cosine} \\
$\gamma$ & PPO/SAC & 0.99 \\
Device & PPO/SAC & CPU \\
Activation function & PPO/SAC & ReLU \\
\hline
\multicolumn{3}{c}{\textbf{PPO}} \\
\hline
$n_{\text{steps}}$ & PPO & 2048 \\
$n_{\text{epochs}}$ & PPO & 4 \\
Batch size & PPO & 512 \\
Clip range, $\varepsilon$  & PPO & 0.15 \\
GAE, $\lambda$ & PPO & 0.95 \\
Value function coefficient, $c_1$ & PPO & 0.5 \\
Entropy coefficient, $c_2$  & PPO & 0.005 \\
Target KL divergence & PPO & 0.15 \\
Maximum gradient norm & PPO & 0.3 \\
\hline
\multicolumn{3}{c}{\textbf{SAC}} \\
\hline
Batch size & SAC & 256 \\
Buffer size & SAC & $10^6$ \\
Learning starts & SAC & $10^4$ \\
Target smoothing $\tau$ & SAC & 0.02 \\
Training frequency (steps) & SAC & Every step \\
Gradient steps per update & SAC & 1 \\
Entropy coefficient, $\alpha$ & SAC & \texttt{auto} \\
Target update interval (steps) & SAC & 1 \\
\hline
\end{tabular}
\end{table}

\subsection{Learning Feasibility}
\begin{figure}[htbp]
    \centering
    \includegraphics[width=0.45\textwidth]{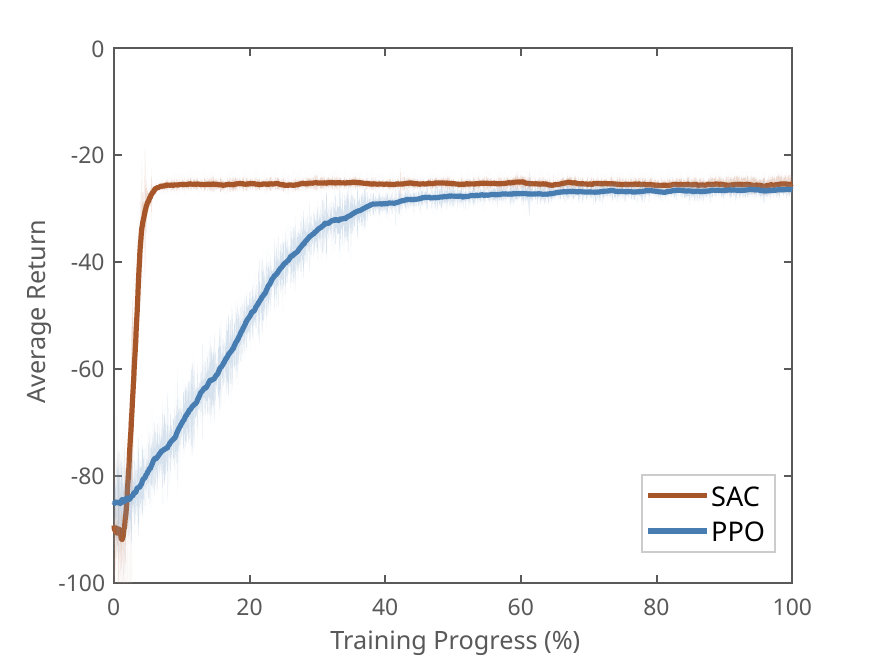}
    \caption{Training performance of \ac{PPO} and \ac{SAC} algorithms under high-latency conditions.  The horizontal axis corresponds to the training progress (\%), and the vertical axis corresponds to the average return.}
    \label{fig:ppo_sac_high_latency}
\end{figure}

\begin{figure}[htbp]
    \centering
    \includegraphics[width=0.45\textwidth]{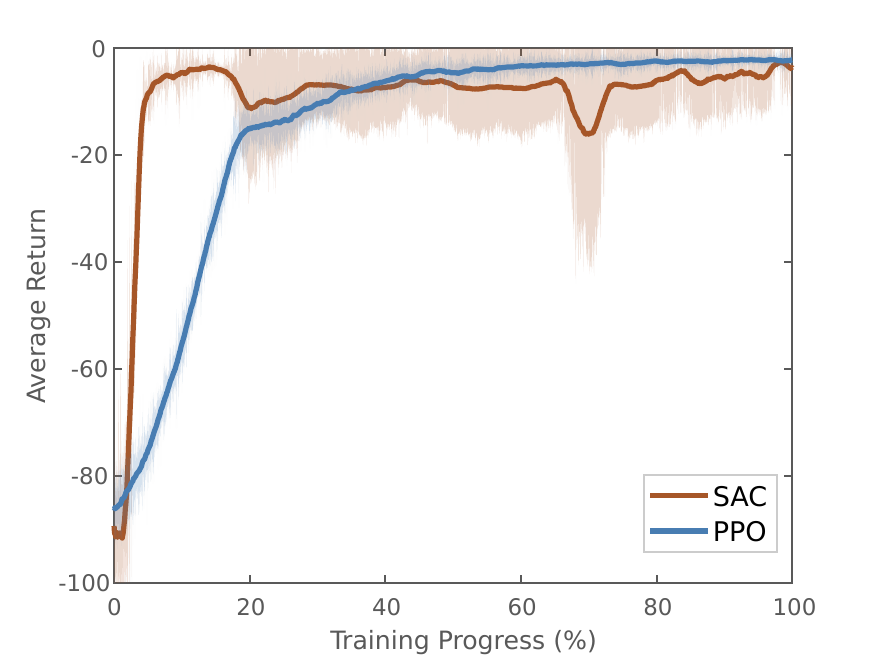}
    \caption{Training performance of \ac{PPO} and SAC algorithms under low latency conditions.}
    \label{fig:ppo_sac_low_latency}
\end{figure}

\noindent
Fig.~\ref{fig:ppo_sac_high_latency} shows the reward curves for \ac{PPO} and SAC for scenarios with large \ac{V2X} communication delays, while Fig.~\ref{fig:ppo_sac_low_latency} depicts the corresponding reward curves for small delays. 
The solid lines denote the mean average return, and the shaded areas show the standard deviation over multiple training tests with different random seeds.

In the high-delay case, both algorithms achieve steady convergence in the average return, with the final return values being nearly identical across different delay settings. This can be attributed to the fact that under high \ac{V2X} latency, short available braking time limits the controllable window for policy. This results in more severe collisions and therefore limits the room for performance improvements through learning.

When the time delay is small, \ac{SAC} exhibits a faster convergence than \ac{PPO}, whereas \ac{PPO} exhibits more stable convergence than \ac{SAC}. Both \ac{DRL} algorithms achieve higher returns than in the high-latency case, indicating improved safety under low-delay conditions.
This improved safety performance for small time delays is mainly due to an extended time window available for decisions.

The training curves show higher variance when the delay is low. This mostly stems from the sparse-reward design: as more controllable scenarios are introduced, positive rewards occur intermittently and in bursts, which amplifies fluctuations in returns and, consequently, in the policy updates~\cite{NgReward} \cite{zhangReward}.

From an application perspective, these results imply that \ac{PPO} may be a more suitable choice for deployment in high-latency scenarios where stability is critical, while \ac{SAC} is to be preferred in low-latency environments where faster convergence is desired.

\begin{figure}[htbp]
    \centering
    \includegraphics[width=0.45\textwidth]{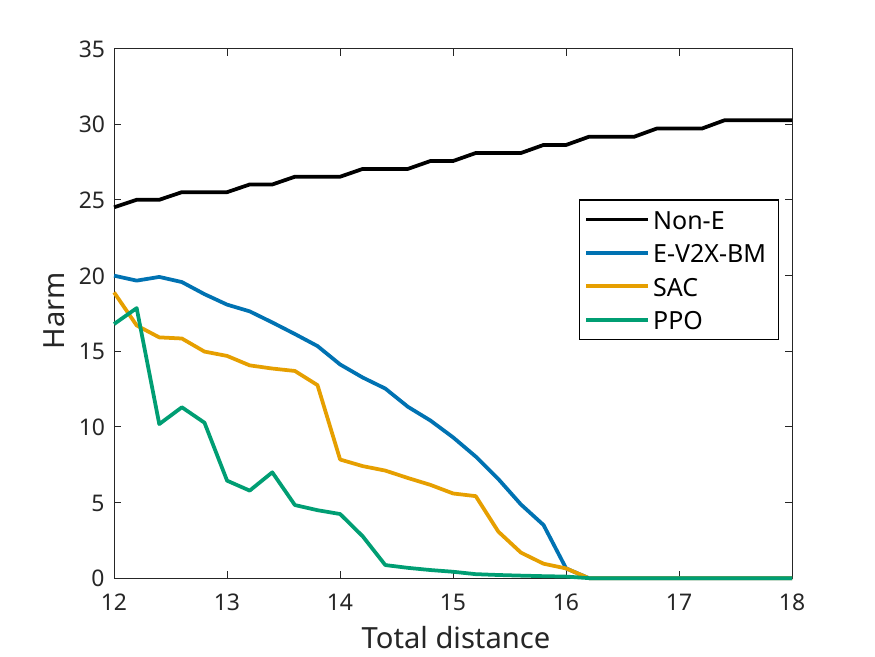}
    \caption{Comparison of harm across four control strategies: Non-ethical, 
    \ac{E-V2X-BM}, and the \ac{DRL} methods \ac{SAC} and \ac{PPO}. 
    The horizontal axis shows the sum of initial \acp{IVD}, i.e. $d_1(0)+d_2(0)$, referred to as total distance, and the vertical axis shows harm.
    \label{Fig:Harm-ivds}}
\end{figure}

\begin{figure}[htbp] 
    \centering
    \includegraphics[width=0.45\textwidth]{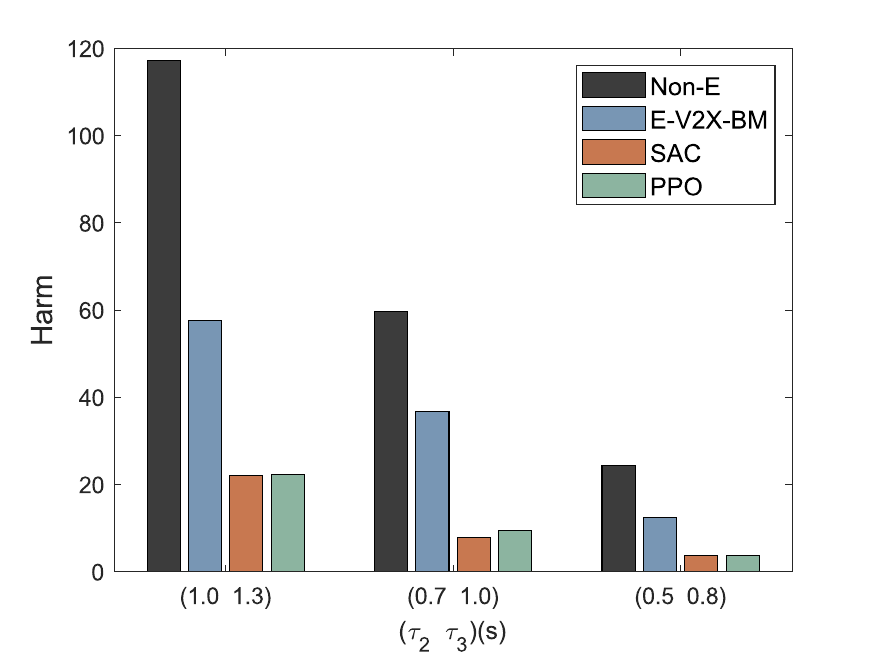}
    \caption{Harm comparison for four control strategies under different \ac{V2X} delays, $(\tau_2, \tau_3)$.}
    \label{Fig:harm-delay}
\end{figure}
\begin{figure*}[!t] 
    \centering
    \includegraphics[width=1\textwidth]{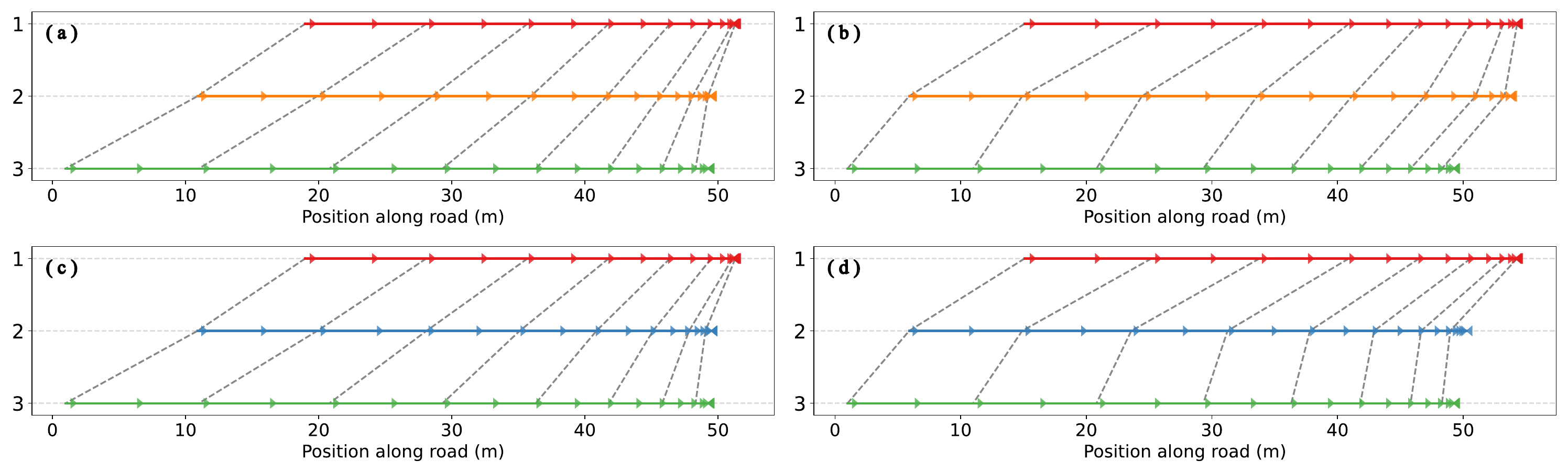}
    \caption{Vehicle trajectories for two scenarios, Scenario 1 and Scenario 2. For each subplot, the x-axis shows the position along the road in meters, and the y-axis denotes the vehicle index. (a) and (b) correspond to PPO in Scenario~1 and 2, respectively, while (c) and (d) correspond to SAC in Scenario~1 and 2, respectively. \label{Fig:pic5sim}}
\end{figure*}
\subsection{Standalone DRL}
For the safety performance evaluation, we compared the severity of collisions (harm) to two different control strategies. The first is a non-ethical approach, where each vehicle applies maximum braking until standstill or collision at the times $0$, $\tau_2$ and $\tau_3$ for Vehicle~1, Vehicle~2 and Vehicle~3, respectively. 
The second is the \ac{E-V2X-BM} method. The two methods provide reference points for evaluating the two standalone (i.e., not the hybrid method) \ac{DRL} strategies, i.e.,  \ac{SAC} respective \ac{PPO}.

In Fig.~\ref{Fig:Harm-ivds} a comparison is shown of the collision severity or harm as function of $d_1(0) + d_2(0)$, referred to as total distance, for the four control strategies when $v_1(0) = 20$~m/s, $v_2(0)=18$~m/s, $v_3(0)= 20$~m/s, $\tau_2=0.5$~s, $\tau_3= 0.8$~s, $a_{l,1}^{\max}=a_{u,1}^{\max} = a_{l,3}^{\max} = a_{u,3}^{\max} = 6~m/s^2$, $a_{l,2}^{\max}=a_{u,2}^{\max}=7~m/s^2$, $m_1 = 4.5$~t, $m_2 = 5.5$~t, $m_3 = 5.9$~t. Here random IVDs are selected for each choice of total distance and average harm is presented. 
The non-ethical strategy has the highest harm, remaining large throughout \acp{IVD} sum test range.  The perhaps non-intuitive result for the non-ethical strategy that larger \acp{IVD} correspond to larger average harm is due to the fact that $a_{l,2}^{\max} > a_{l,3}^{\max}$. Since Vehicle~2 brakes earlier and harder than the Vehicle~3, an initial \ac{IVD} $d_2(0)$ allows (statistically speaking) the relative velocity between Vehicle 2 and Vehicle 3 to be larger at collision.

The \ac{E-V2X-BM} strategy reduces harm compared to the non-ethical method, but still suffers from high collision severity for short initial \acp{IVD}, reflecting the limitations of selecting a constant deceleration for Vehicle 2. 
Both \ac{SAC} and \ac{PPO} strategies achieve lower average harm compared to the \ac{E-V2X-BM} strategy across the entire \acp{IVD} range.

Fig.~\ref{Fig:harm-delay} presents harm values for the four strategies under three different communication delay configurations, where the initial \acp{IVD} are $d_1(0)=d_2(0)=7$~m, and all other settings as in the previous scenario.
Both \ac{SAC} and \ac{PPO} consistently achieve the lowest harm values across all tested delay setting, demonstrating strong robustness to increased communication latency.  


To further examine the detailed operational behavior of the control strategies, simulation experiments are performed under two sets of initial conditions. 
Figure~\ref{Fig:pic5sim} presents the resulting vehicle-following trajectories for \ac{PPO} and \ac{SAC}. 
For clarity, the positions of the three vehicles along the road are illustrated as parallel lines, with vehicle headings indicated by arrows at every tenth time step. Dashed lines connect the positions of the vehicles for the same time steps, showing their relative spacing.

For subplots~(a) and~(c), the initial \acp{IVD} are \(d_1(0) = 10\ \mathrm{m}\) and \(d_2(0) = 8\ \mathrm{m}\), and the initial velocities are \(v_1(0) = 20\ \mathrm{m/s}\), \(v_2(0) = 18\ \mathrm{m/s}\), and \(v_3(0) = 20\ \mathrm{m/s}\). 
For subplots~(b) and~(d), the corresponding initial values are \(d_1(0) = 5\ \mathrm{m}\), \(d_2(0) = 9\ \mathrm{m}\), \(v_1(0) = 22\ \mathrm{m/s}\), \(v_2(0) = 18\ \mathrm{m/s}\), and \(v_3(0) = 20\ \mathrm{m/s}\). 

In Fig.~\ref{Fig:pic5sim}, it can be observed that vehicle positions are similar for \ac{SAC} and \ac{PPO} in Scenario~1. However, in Scenario~2, despite identical reward shaping and parameter settings, the two \ac{DRL} algorithms result in noticeably different positioning. For Vehicle~2, \ac{PPO} provides a policy that prioritizes maintaining a larger distance to Vehicle~3, while \ac{SAC} provides a policy that prioritizes maintaining a larger distance to Vehicle~1. 


\begin{figure}[!t] 
    \centering
    \includegraphics[width=0.5\textwidth]{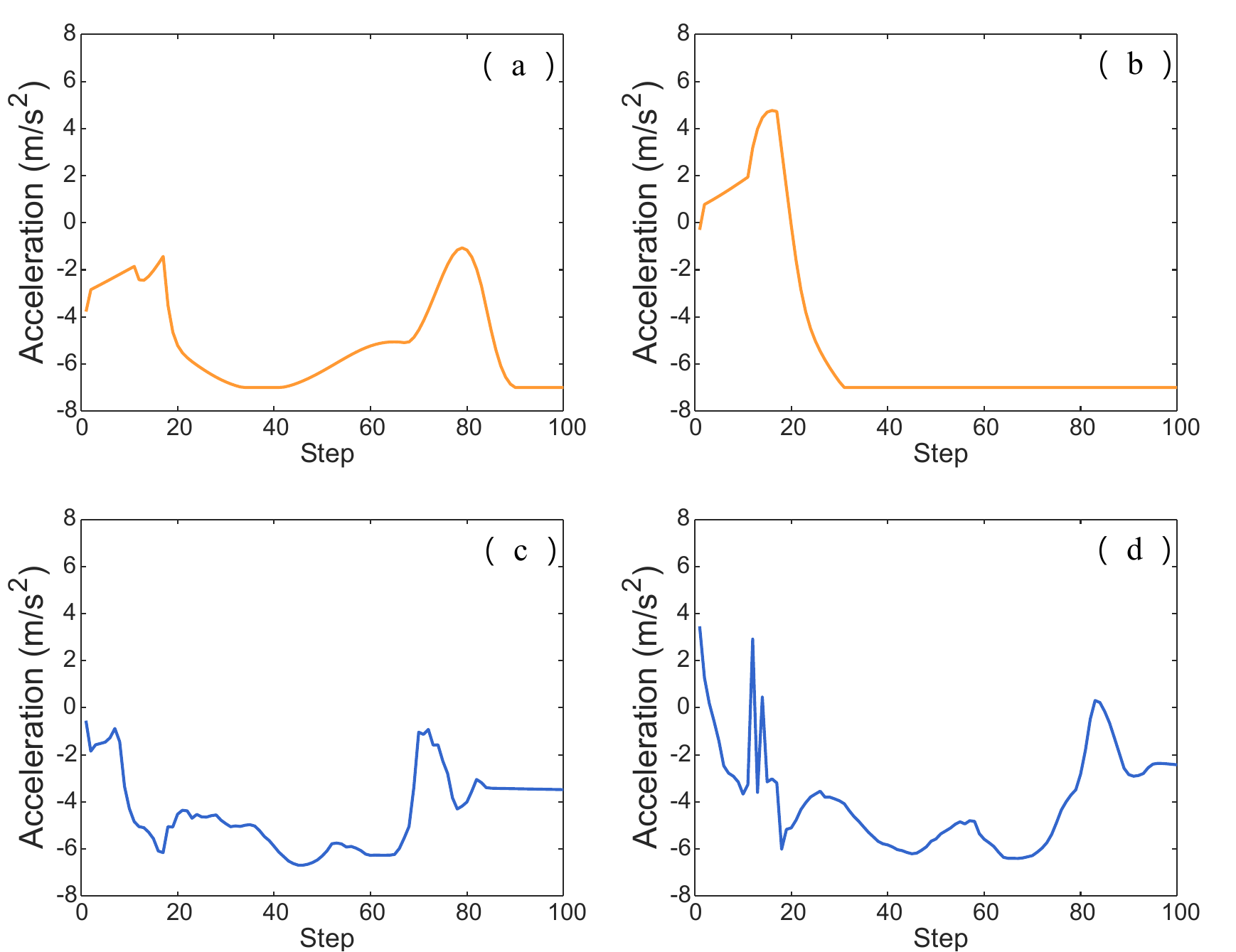}
    \caption {
    Acceleration/deceleration profiles under two different scenarios. 
    (a) PPO, Scenario 1. 
    (b) PPO, Scenario 2. 
    (c) SAC, Scenario 1. 
    (d) SAC, Scenario 2. 
    \label{Fig:pic5act}
    }
\end{figure}

Fig.~\ref{Fig:pic5act} shows the control strategies obtained from \ac{PPO} and \ac{SAC} that produce the vehicle positions in Fig.\ref{Fig:pic5sim}. Subplots~(a) and~(c) in Fig.~\ref{Fig:pic5act} show similar patterns, which was also observed for the corresponding subplots (a) and (c) in Fig.~\ref{Fig:pic5sim}. In contrast, subplots~(b) and~(d), corresponding to (b) and (d) in Fig.~\ref{Fig:pic5sim}, show distinct behaviors for Scenario~2; for \ac{PPO} the control strategy involves an initial acceleration prior to maximum braking, whereas \ac{SAC} exhibits deceleration throughout most of the scenario. For scenario 2, despite being trained with the same reward shaping, the \ac{PPO} policy produces a smoother acceleration/deceleration profile compared to \ac{SAC}.

\subsection{Hybrid Method}

\begin{figure}[!t] 
    \centering
    \includegraphics[width=0.45\textwidth]{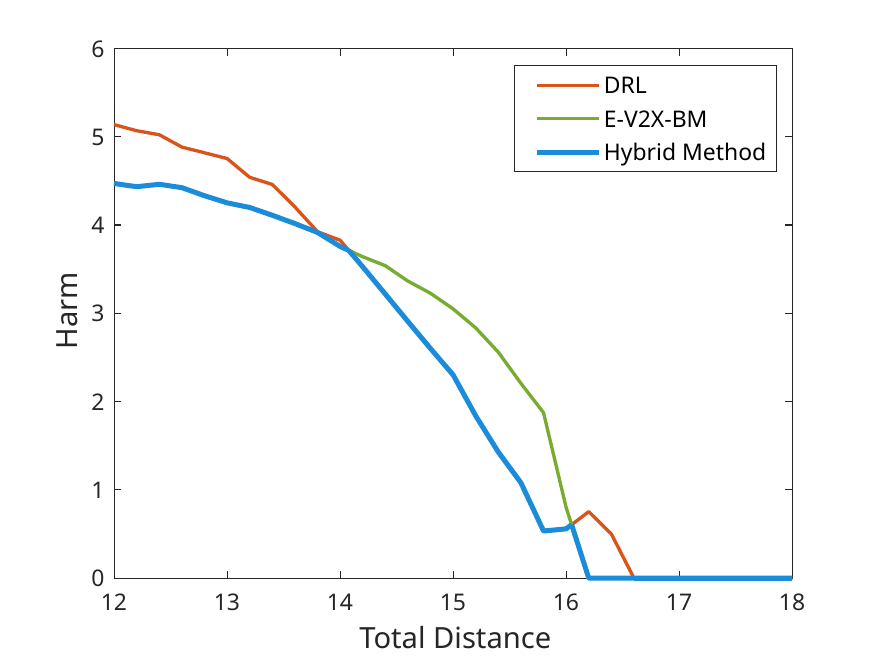}
    \caption 
    {\label{Fig:hybrid illu}
    Illustration of the hybrid method. Parameters for simulation are the same as for Fig.\ref{Fig:Harm-ivds}.
    }
\end{figure}

This section presents an evaluation of the proposed hybrid approach that combines \ac{DRL} with \ac{E-V2X-BM}.
Figure~\ref{Fig:hybrid illu} captures the concept of the hybrid approach. The green curve is the harm for \ac{E-V2X-BM} as a function of total distance (i.e., $d_1(0) + d_2(0)$), while the orange curve is the harm for the standalone \ac{DRL} model as function of total distance. Here, to capture potential suboptimal \ac{DRL} strategies that may occur in practice, we considered a \ac{PPO} model obtained from training where convergence has not occurred. According to Algorithm~\ref{alg:hybrid}, the hybrid approach consistently selects the strategy that results in lower harm, which is depicted in blue. 

To further validate the effectiveness of the proposed hybrid method, we conduct comparative experiments with standalone \ac{DRL} and \ac{E-V2X-BM} strategies, using identical simulation settings. Performance is measured under varying \ac{IVD} conditions ($d_1(0), d_2(0)$ in the range $[5,10]$ m) over 5,000 test episodes with a fixed random seed. To highlight the importance of incorporating a safety bound, we considered a \ac{PPO} model obtained from training before convergence occurred. 
\small
\begin{table}[!t]
    \centering 
    \caption{Performance of \ac{E-V2X-BM}, \ac{DRL}, and hybrid strategies}\label{Table:hybrid}
    \renewcommand{\arraystretch}{1.5} 
    \begin{tabular}{l c c c }
        \toprule
        \textbf{Metric}  & \textbf{E-V2X-BM} & \textbf{Standalone DRL} & \textbf{Hybrid method} \\
        \midrule
         Number of   & 4263 & 3737 & 2995 \\
         collisions  &   &   &  \\
         Collision rate   & 85.26\% & 74.74\% & 59.9\% \\
        Average Harm     &9.8514 & 12.229 & 7.3641 \\
        \bottomrule
    \end{tabular}
\end{table}
\normalsize 

The results are summarized in Table \ref{Table:hybrid}. The hybrid method achieves the lowest number of average harm, collisions, and collision rate (which is the fraction of scenarios ending in collisions). It is  outperforming both the \ac{E-V2X-BM}-only baseline and the standalone \ac{DRL} model. 

To further demonstrate the effectiveness of \ac{DRL} in mitigating collision severity, we considered scenarios where $d_{1}(0)$ and $d_{2}(0)$ were randomly sampled in the interval $[5, 10]$~m, and $v_{1}(0),v_{2}(0),v_{3}(0)$ were randomly sampled in the interval $[18, 22]~\mathrm{m/s}$. This setting was chosen to examine scenarios with small \acp{IVD} and moderate-speed vehicle-following, where the advantages of \ac{DRL} are expected to be most pronounced. All other parameters are identical to those in Figure~\ref{Fig:Harm-ivds}. 
A total of 10{,}000 random scenarios were generated. In each scenario, the performance of the non-ethical method (denoted as Non-E), \ac{E-V2X-BM}, and the hybrid methods based on \ac{SAC} and \ac{PPO} was evaluated. For each method, the number of collisions, collision rate, and the average harm were computed, followed by calculation of the reduction in harm relative to the non-ethical method. The results are summarized in Table~\ref{Table:Random}.

\begin{table}[!t]
    \centering 
    \caption{Comparison of four different strategies (random test)}\label{Table:Random}
    \renewcommand{\arraystretch}{1.5} 
    \begin{tabular}{l c c c c}
        \toprule
        \textbf{Metric} & \textbf{Non-E} & \textbf{E-V2X-BM} & \textbf{SAC} & \textbf{PPO} \\
        \midrule
        Number of collisions & 9391 & 8520 & 5896 & 5974 \\
        Collision rate & 93.91\% & 85.2\% & 58.96\% & 59.74\% \\
        Average Harm  & 33.5878 & 9.8772 & 5.7001 & 5.6202 \\
        Harm decrease  & 0\% & 70.59\% & 83.02\% & 83.26\% \\
        \bottomrule
    \end{tabular}
\end{table}

The non-ethical method has the highest number of collision episodes ($9391$) and collision rate ($93.91\%$), along with the largest average harm ($33.5878$). The \ac{E-V2X-BM} method reduces both collision number and average harm, achieving a harm decrease ratio of $70.59\%$ relatively to non-ethical method. Hybrid methods based on \ac{SAC} and \ac{PPO} demonstrate the best overall performance, with collision rates of $58.96\%$ and $59.74\%$, respectively, and the lowest average harm values ($5.7001$ for \ac{SAC} and $5.6202$ for \ac{PPO}). Both hybrid methods, based on \ac{SAC} and \ac{PPO}, achieve over $83\%$ reduction in harm compared with the non-ethical method, indicating \ac{DRL} superior capability in mitigating collisions and minimizing severity.



\section{Conclusions}
\label{sec:conclusion}
The paper demonstrates that \acf{DRL} can significantly enhance safety in multi-vehicle-following scenarios. Two algorithms, \acf{PPO} and \acf{SAC}, were evaluated for deriving effective emergency braking strategies. Furthermore, a hybrid approach was proposed to improve the reliability and robustness of the \ac{DRL}-based policies by integrating them with a conservative safety bound. Results show that combining \ac{DRL} with the safety bound significantly reduces the collision rate and average collision severity/harm in situations with potential collision risks. Overall, the proposed hybrid method achieves superior safety and demonstrates potential for practical deployment.



\bibliographystyle{IEEEtran}
\bibliography{references}

\begin{thebibliography}{10}
\providecommand{\url}[1]{#1}
\csname url@samestyle\endcsname
\providecommand{\newblock}{\relax}
\providecommand{\bibinfo}[2]{#2}
\providecommand{\BIBentrySTDinterwordspacing}{\spaceskip=0pt\relax}
\providecommand{\BIBentryALTinterwordstretchfactor}{4}
\providecommand{\BIBentryALTinterwordspacing}{\spaceskip=\fontdimen2\font plus
\BIBentryALTinterwordstretchfactor\fontdimen3\font minus \fontdimen4\font\relax}
\providecommand{\BIBforeignlanguage}[2]{{%
\expandafter\ifx\csname l@#1\endcsname\relax
\typeout{** WARNING: IEEEtran.bst: No hyphenation pattern has been}%
\typeout{** loaded for the language `#1'. Using the pattern for}%
\typeout{** the default language instead.}%
\else
\language=\csname l@#1\endcsname
\fi
#2}}
\providecommand{\BIBdecl}{\relax}
\BIBdecl

\bibitem{AVtraffic}
P.~Ghorai, A.~Eskandarian, M.~Abbas, and A.~Nayak, ``A causation analysis of autonomous vehicle crashes,'' \emph{IEEE Intelligent Transportation Systems Magazine}, vol.~16, no.~5, pp. 33--45, 2024.

\bibitem{TOURAN1999567}
A.~Touran, M.~A. Brackstone, and M.~McDonald, ``A collision model for safety evaluation of autonomous intelligent cruise control,'' \emph{Accident Analysis \& Prevention}, vol.~31, no.~5, pp. 567--578, 1999.

\bibitem{rearend1}
I.~Taourarti, A.~Ramaswamy, J.~Ibanez-Guzman, B.~Monsuez, and A.~Tapus, ``Cross-cultural analysis of car-following dynamics: A comparative study of open-source trajectory datasets,'' in \emph{2025 IEEE Intelligent Vehicles Symposium (IV)}, 2025, pp. 330--337.

\bibitem{sidorenko2023ethical}
G.~Sidorenko, J.~Thunberg, and A.~Vinel, ``Ethical v2x: Cooperative driving as the only ethical path to multi-vehicle safety,'' in \emph{2023 IEEE 98th Vehicular Technology Conference (VTC2023-Fall)}.\hskip 1em plus 0.5em minus 0.4em\relax IEEE, 2023, pp. 1--6.

\bibitem{geisslinger2023ethical}
M.~Geisslinger, F.~Poszler, and M.~Lienkamp, ``An ethical trajectory planning algorithm for autonomous vehicles,'' \emph{Nature Machine Intelligence}, vol.~5, no.~2, pp. 137--144, 2023.

\bibitem{katrakazas2015real}
C.~Katrakazas, M.~Quddus, W.-H. Chen, and L.~Deka, ``Real-time motion planning methods for autonomous on-road driving: State-of-the-art and future research directions,'' \emph{Transportation Research Part C: Emerging Technologies}, vol.~60, pp. 416--442, 2015.

\bibitem{sidorenko2024cooperation}
G.~Sidorenko, J.~Thunberg, and A.~Vinel, ``Cooperation for ethical autonomous driving,'' in \emph{2024 20th International Conference on Wireless and Mobile Computing, Networking and Communications (WiMob)}.\hskip 1em plus 0.5em minus 0.4em\relax IEEE, 2024, pp. 391--395.

\bibitem{DRL1}
T.~P. Lillicrap, J.~J. Hunt, A.~Pritzel, N.~Heess, T.~Erez, Y.~Tassa, D.~Silver, and D.~Wierstra, ``Continuous control with deep reinforcement learning,'' \emph{arXiv preprint arXiv:1509.02971}, 2015.

\bibitem{DRL2}
V.~Mnih, K.~Kavukcuoglu, D.~Silver, A.~A. Rusu, J.~Veness, M.~G. Bellemare, A.~Graves, M.~A. Riedmiller, A.~K. Fidjeland, G.~Ostrovski, S.~Petersen, C.~Beattie, A.~Sadik, I.~Antonoglou, H.~King, D.~Kumaran, D.~Wierstra, S.~Legg, and D.~Hassabis, ``Human-level control through deep reinforcement learning,'' \emph{Nature}, vol. 518, pp. 529--533, 2015.

\bibitem{AVDRL}
S.~Aradi, ``Survey of deep reinforcement learning for motion planning of autonomous vehicles,'' \emph{IEEE Transactions on Intelligent Transportation Systems}, vol.~23, no.~2, pp. 740--759, 2022.

\bibitem{DRLcontrol1}
T.~P. Lillicrap, J.~J. Hunt, A.~Pritzel, N.~Heess, T.~Erez, Y.~Tassa, D.~Silver, and D.~Wierstra, ``Continuous control with deep reinforcement learning,'' \emph{arXiv preprint arXiv:1509.02971}, 2015.

\bibitem{DRLcar1}
J.~L. Vermeulen, A.~Hillebrand, and R.~Geraerts, ``Annotating traversable gaps in walkable environments,'' in \emph{2018 IEEE International Conference on Robotics and Automation (ICRA)}, 2018, pp. 3045--3052.

\bibitem{DRLcontrol2}
A.~Kendall, J.~Hawke, D.~Janz, P.~Mazur, D.~Reda, J.-M. Allen, V.-D. Lam, A.~Bewley, and A.~Shah, ``Learning to drive in a day,'' in \emph{2019 international conference on robotics and automation (ICRA)}.\hskip 1em plus 0.5em minus 0.4em\relax IEEE, 2019, pp. 8248--8254.

\bibitem{DRLcar2}
R.~Eshleman and R.~Singh, ``Reconstructing the temporal progression of biological data using cluster spanning trees,'' \emph{IEEE Transactions on NanoBioscience}, vol.~16, no.~2, pp. 140--147, 2017.

\bibitem{longitudinal}
I.~Taourarti, A.~Ramaswamy, J.~Ibanez-Guzman, B.~Monsuez, and A.~Tapus, ``Cross-cultural analysis of car-following dynamics: A comparative study of open-source trajectory datasets,'' in \emph{2025 IEEE Intelligent Vehicles Symposium (IV)}, 2025, pp. 330--337.

\bibitem{longitudinal2}
P.~F. Orzechowski, K.~Li, and M.~Lauer, ``Towards responsibility-sensitive safety of automated vehicles with reachable set analysis,'' in \emph{2019 IEEE International Conference on Connected Vehicles and Expo (ICCVE)}, 2019, pp. 1--6.

\bibitem{rearend}
L.~Li, G.~Lu, Y.~Wang, and D.~Tian, ``A rear-end collision avoidance system of connected vehicles,'' in \emph{17th International IEEE Conference on Intelligent Transportation Systems (ITSC)}, 2014, pp. 63--68.

\bibitem{V2X}
S.~Chen, J.~Hu, Y.~Shi, Y.~Peng, J.~Fang, R.~Zhao, and L.~Zhao, ``Vehicle-to-everything (v2x) services supported by lte-based systems and 5g,'' \emph{IEEE Communications Standards Magazine}, vol.~1, no.~2, pp. 70--76, 2017.

\bibitem{Ethical}
Y.~Wang, G.~Tan, and H.~Si, ``Decision modeling for automated driving in dilemmas based on bidirectional value alignment of moral theory values and fair human moral values,'' \emph{Transportation Research Part F: Traffic Psychology and Behaviour}, vol. 108, pp. 14--27, 2025.

\bibitem{Harm}
D.~J. Gabauer and H.~C. Gabler, ``Comparison of roadside crash injury metrics using event data recorders,'' \emph{Accident Analysis \& Prevention}, vol.~40, no.~2, pp. 548--558, 2008.

\bibitem{Harm2}
J.~Zhang, C.~Liang, S.~Yu, R.~Liu, and J.~Gao, ``Consistency of calculation results of two typical vehicle collision models,'' \emph{Procedia Engineering}, vol. 137, pp. 220--224, 2016, green Intelligent Transportation System and Safety.

\bibitem{Harm3}
T.~Semple and G.~Fountas, ``Twelve years of evidence: modelling the injury severity of single-vehicle collisions pre- and post-20mph (32 km/h) implementation in edinburgh and glasgow,'' \emph{Accident Analysis \& Prevention}, vol. 221, p. 108183, 2025.

\bibitem{Harm-e}
A.~F. Tencer, ``Low speed rear end automobile collisions and whiplash injury, the biomechanical approach,'' \emph{Medicine, Law \& Society}, vol.~12, no.~2, p. 1–20, Oct. 2019.

\bibitem{AVDRL2}
L.~Crosato, H.~P.~H. Shum, E.~S.~L. Ho, and C.~Wei, ``Interaction-aware decision-making for automated vehicles using social value orientation,'' \emph{IEEE Transactions on Intelligent Vehicles}, vol.~8, no.~2, pp. 1339--1349, 2023.

\bibitem{DRLdriving}
C.~Yu, X.~Wang, X.~Xu, M.~Zhang, H.~Ge, J.~Ren, L.~Sun, B.~Chen, and G.~Tan, ``Distributed multiagent coordinated learning for autonomous driving in highways based on dynamic coordination graphs,'' \emph{IEEE Transactions on Intelligent Transportation Systems}, vol.~21, no.~2, pp. 735--748, 2020.

\bibitem{HardwareDRL2}
A.~Kendall, J.~Hawke, D.~Janz, P.~Mazur, D.~Reda, J.-M. Allen, V.-D. Lam, A.~Bewley, and A.~Shah, ``Learning to drive in a day,'' in \emph{2019 international conference on robotics and automation (ICRA)}.\hskip 1em plus 0.5em minus 0.4em\relax IEEE, 2019, pp. 8248--8254.

\bibitem{hardwareDRL1}
B.~R. Kiran, I.~Sobh, V.~Talpaert, P.~Mannion, A.~A. Al~Sallab, S.~Yogamani, and P.~P{\'e}rez, ``Deep reinforcement learning for autonomous driving: A survey,'' \emph{IEEE transactions on intelligent transportation systems}, vol.~23, no.~6, pp. 4909--4926, 2021.

\bibitem{HardwareDRLdelay}
A.~Hazra, V.~M.~R. Tummala, N.~Mazumdar, D.~K. Sah, and M.~Adhikari, ``Deep reinforcement learning in edge networks: Challenges and future directions,'' \emph{Physical Communication}, vol.~66, p. 102460, 2024.

\bibitem{DRLblackbox}
E.~Puiutta and E.~M. Veith, ``Explainable reinforcement learning: A survey,'' in \emph{International cross-domain conference for machine learning and knowledge extraction}.\hskip 1em plus 0.5em minus 0.4em\relax Springer, 2020, pp. 77--95.

\bibitem{PPO}
J.~Schulman, F.~Wolski, P.~Dhariwal, A.~Radford, and O.~Klimov, ``Proximal policy optimization algorithms,'' \emph{arXiv preprint arXiv:1707.06347}, 2017.

\bibitem{SAC1}
T.~Haarnoja, A.~Zhou, P.~Abbeel, and S.~Levine, ``Soft actor-critic: Off-policy maximum entropy deep reinforcement learning with a stochastic actor,'' in \emph{International conference on machine learning}.\hskip 1em plus 0.5em minus 0.4em\relax Pmlr, 2018, pp. 1861--1870.

\bibitem{SAC2}
T.~Haarnoja, A.~Zhou, K.~Hartikainen, G.~Tucker, S.~Ha, J.~Tan, V.~Kumar, H.~Zhu, A.~Gupta, P.~Abbeel \emph{et~al.}, ``Soft actor-critic algorithms and applications,'' \emph{arXiv preprint arXiv:1812.05905}, 2018.

\bibitem{Sigmoid}
D.~Silver, R.~Sutton, and M.~M\"{u}ller, ``Reinforcement learning of local shape in the game of go,'' in \emph{Proceedings of the 20th International Joint Conference on Artifical Intelligence}, ser. IJCAI'07.\hskip 1em plus 0.5em minus 0.4em\relax San Francisco, CA, USA: Morgan Kaufmann Publishers Inc., 2007, p. 1053–1058.

\bibitem{NgReward}
A.~Y. Ng, D.~Harada, and S.~J. Russell, ``Policy invariance under reward transformations: Theory and application to reward shaping,'' in \emph{Proceedings of the Sixteenth International Conference on Machine Learning}, ser. ICML '99.\hskip 1em plus 0.5em minus 0.4em\relax San Francisco, CA, USA: Morgan Kaufmann Publishers Inc., 1999, p. 278–287.

\bibitem{zhangReward}
S.~Zhang and R.~S. Sutton, ``A deeper look at experience replay,'' \emph{arXiv preprint arXiv:1712.01275}, 2017.

\end{thebibliography}

\end{document}